\documentclass{elsarticle}

\usepackage{multicol}
\usepackage{color}
\usepackage{xspace}
\usepackage{enumerate}
\usepackage{url}
\usepackage[frozencache=true,cachedir=minted-cache]{minted} 
\usepackage{booktabs}

\usepackage{pdflscape}
\usepackage{array}
\newcolumntype{C}[1]{>{\centering\arraybackslash}p{#1}}
\newcolumntype{R}[1]{>{\raggedleft\arraybackslash}p{#1}}
\newcolumntype{L}[1]{>{\raggedright\arraybackslash}p{#1}}
\usepackage{amsmath}
\usepackage{multirow}

\def\ttdefault{cmtt}

\makeatletter
\def\bs{\expandafter\@gobble\string\\}
\def\lb{\expandafter\@gobble\string\{}
\def\rb{\expandafter\@gobble\string\}}
\def\@pdfauthor{J.F.Silva}
\def\@pdftitle{Modelling Patient Trajectories Using Multimodal Information}
\def\@pdfsubject{Research article on patient trajectory modelling using multimodal clinical information.}
\def\@pdfkeywords{Patient Trajectory Modelling, Deep Learning, Contextual Embeddings, Clinical Notes, EHR}

\DeclareRobustCommand{\LaTeX}{L\kern-.26em%
        {\sbox\z@ T%
         \vbox to\ht\z@{\hbox{\check@mathfonts
           \fontsize\sf@size\z@
           \math@fontsfalse\selectfont
          A\,}%
         \vss}%
        }%
     \kern-.15em%
    \TeX}
\makeatother

\patchcmd{\pprintMaketitle}
 {\hrule}
 {\clearpage\hrule}
 {}{}
\appto\endfrontmatter{\clearpage}

\begin{document}

\begin{frontmatter}
\title{Modelling Patient Trajectories Using Multimodal Information}

\author[1]{João Figueira Silva\corref{cor1}}
\ead{joaofsilva@ua.pt}

\author[1]{Sérgio Matos}
\ead{aleixomatos@ua.pt}

\cortext[cor1]{Corresponding author}
\address[1]{DETI/IEETA, University of Aveiro, Aveiro, Portugal}

\date{May 17, 2022}

\pagebreak

\begin{abstract}
  
  \textbf{Background:} Electronic Health Records~(EHRs) aggregate diverse information at the patient level, holding a trajectory representative of the evolution of the patient health status throughout time. Although this information provides context and can be leveraged by physicians to monitor patient health and make more accurate prognoses/diagnoses, patient records can contain information from very long time spans, which combined with the rapid generation rate of medical data makes clinical decision making more complex. Patient trajectory modelling can assist by exploring existing information in a scalable manner, and can contribute in augmenting health care quality by fostering preventive medicine practices (\textit{e.g.} earlier disease diagnosis).
  
  \textbf{Methods:} We propose a solution to model patient trajectories that combines different types of information (\textit{e.g.} clinical text, standard codes) 
  and considers the temporal aspect of clinical data. This solution leverages two different architectures: one supporting flexible sets of input features, to convert patient admissions into dense representations; and a second exploring extracted admission representations in a recurrent-based architecture, where patient trajectories are processed in sub-sequences using a sliding window mechanism.
  
  \textbf{Results:} The developed solution was evaluated on two different clinical outcomes, unexpected patient readmission and disease progression, using the publicly available Medical Information Mart for Intensive Care~(MIMIC)-III clinical database. The results obtained demonstrate the potential of the first architecture to model readmission and diagnoses prediction using single patient admissions. While information from clinical text did not show the discriminative power observed in other existing works, this may be explained by the need to fine-tune the clinicalBERT model. Finally, we demonstrate the potential of the sequence-based architecture using a sliding window mechanism to represent the input data, attaining comparable performances to other existing solutions.
 
  \textbf{Conclusion:} 
  
  Herein, we explored DL-based techniques to model patient trajectories and propose two flexible architectures that explore patient admissions on an individual and sequence basis. The combination of clinical text with other types of information led to positive results, which can be further improved by including a fine-tuned version of clinicalBERT in the architectures. The proposed solution can be publicly accessed at \url{https://github.com/bioinformatics-ua/PatientTM}.
  
\end{abstract}


\end{frontmatter}

\pagebreak

\section{Introduction}

Electronic Health Records~(EHRs) aggregate a multitude of information at the patient level, holding a longitudinal view of the patient medical history and thus containing a trajectory that is representative of the evolution of the patient health status throughout time. This information is important as it provides context and can be used by physicians to monitor patient health and make more accurate prognoses or diagnoses. However, patient records can contain information from very long time spans (\textit{e.g.} decades), which combined with the rapidly increasing rate at which medical data is being generated makes the process of making clinical decisions more complex.

Since generated data is stored in digital format, supervised or semi-supervised solutions can explore these rich medical data sources and leverage them to derive diagnosis/prognosis predictions with the objective of assisting physicians in the clinical decision making process. Extracting and presenting relevant data in explanatory ways are also very important to physicians, who want to understand the data and the reasoning behind projected decisions. Therefore, condensing relevant information from patient records can be of great value for physicians instead of simply providing them with black-box predictions~\cite{lauritsen2020explainableAI}.

Patient trajectory modelling has been attracting increasing research interest due to its possible contribution in augmenting health care quality, for instance by fostering preventive medicine practices. This is particularly relevant as an earlier disease diagnosis can enable better disease management, earlier intervention and better resource allocation~\cite{li2020behrt}. However, the process of developing these solutions must cope with key challenges such as managing to factor in the temporal aspect embedded in medical data, dealing with knowledge from previous events~\cite{LIGDoctor2020}, and addressing common problems in medical data like the high granularity and sparsity, and low volume and cardinality~\cite{LIGDoctor2020, chen2019MLvsDL}. Furthermore, problem setting is also an important aspect since it can greatly influence data properties, as Intensive Care Unit~(ICU) scenarios assume constant monitoring of several relevant vital signals, whereas chronic disease scenarios are characterized by hospital visits with variable periodicity, depending on the disease state, and therefore offer much sparser data~\cite{chen2019MLvsDL}. 

Patient trajectory modelling solutions can be developed with various prediction objectives in mind, such as disease progression by predicting the diagnoses on a future admission, recommended intervention through predicted future procedures~\cite{pham2017DeepCare}, expected elapsed time until a patient is readmitted to the health care system, among other possible clinical outcomes. Herein, we propose a solution to model patient trajectories that combines different types of information and considers the temporal aspect of clinical data. This solution explores a recurrent architecture that leverages extracted patient admission representations, and was developed using data from the public Medical Information Mart for Intensive Care~(MIMIC)-III clinical database~\cite{Johnson2016} considering the prediction of two different clinical outcomes: patient readmission within a period of 30 days after the patient has been discharged, and disease progression through the prediction of expected diagnoses codes on a future admission. The proposed system is publicly available at \url{https://github.com/bioinformatics-ua/PatientTM}.



\section{Related Work}

Although existing reviews on ML applications to longitudinal EHR data have focused on benchmarks~\cite{bellamy2020evaluating} and on deep representation learning~\cite{Si2021EHRreview}, in~\cite{silva2021review} we have presented and discussed a list of existing solutions for patient trajectory modelling, categorizing approaches based on the following 3 key aspects: (i) used methodology, (ii) explored data types, and (iii) integration of the temporal component from data. The temporal aspect of data was given particular focus as it is inherent to the concept of health status progression, and is also one of the most challenging aspects to integrate in trajectory modelling as evidenced by the wide variety of approaches used to capture the temporal component. In this section, we provide an overview of each solution covering the three aforementioned key aspects as well as each solution's practical application regarding clinical outcomes, and refer the reader to~\cite{silva2021review} for a more in-depth description and analysis of each solution.

Jensen~\textit{et al.}~\cite{Jensen2017} proposed an approach for diagnosis and high-risk prediction, which extracted information from EHR free text and used it to recreate patient trajectories. The authors used three Na\"ive Bayes classifiers to classify entities into entity groups (diseases and symptoms, drugs and medication, and surgical procedures), and compiled language features to capture temporal information and distinguish between real-time events and noise, retrospective events and noise, and real-time and retrospective events. Relevant entities were mined using Frequent Item Set mining, and extracted entity sets were used to reconstruct disease trajectories (\textit{e.g.} symptom $\rightarrow$ disease $\rightarrow$ death) and event trajectories (\textit{e.g.} symptom $\rightarrow$ disease $\rightarrow$ admission $\rightarrow$ drug $\rightarrow$ surgery $\rightarrow$ death). Obtained trajectories were analysed to quantify risk and draw causal conclusions~\cite{Jensen2017}. Paik~\textit{et al.}~\cite{Paik2019Scizophrenia} followed a different approach to map trajectories, exploring structured data such as patient demographics, measurements, and others. The authors identified significant disease associations through a statistical analysis, evaluating the relative association of diagnoses that co-occurred within one year in each patient. The decision to use one-year intervals came from the assumption that admissions with very-long distance intervals were less likely to be related. The resulting temporally aligned disease pairs were mapped into Directed Acyclic Graphs~(DAGs) using a greedy search algorithm, and created trajectories were analysed to uncover possible unknown relations between diseases, resulting in the discovery of a potential unexpected risk of rhabdomyolysis in schizophrenic patients~\cite{Paik2019Scizophrenia}.

In DoctorAI, Choi~\textit{et al.}~\cite{choi2016doctorAI} proposed a solution to predict patient status on the next patient interaction, using a Gated Recurrent Unit~(GRU) based model to predict diagnosis and medication codes along with the estimated elapsed time until the next event. This model leveraged standard nomenclature codes for diagnosis, procedures and medication, and integrated the temporal component through the inclusion of patient visit intervals as input feature~\cite{choi2016doctorAI}. Pham~\textit{et al.}~\cite{pham2017DeepCare} proposed DeepCare as a solution for disease progression modelling, intervention recommendation, readmission prediction and high-risk prediction. DeepCare is a Long Short-Term Memory~(LSTM) model based on modified LSTM units named by the authors as Care LSTM, which model admission methods, irregular timing, diagnosis and intervention information. The authors used structured information and time intervals as model input, and forced a multiscale temporal structure on top of the proposed Care-LSTM that pools historical illness states within multiple time-horizons~\cite{pham2017DeepCare}. Wu~\textit{et al.}~\cite{wu2018Asynchronous} used a LSTM based architecture to model asynchronous medical event sequences and predict disease progression as well as mortality, proposing modified LSTM units that contain additional time gates used to integrate timing information in the modelling process, and evaluating the impact of several strategies for incorporating the temporal component in model performance. This approach extracted structured data from free text using a Natural Language Processing~(NLP) feature extraction component and a set of logic-based rules, and combined it with other previously existing structured data~\cite{wu2018Asynchronous}. 

Though not entirely focused on patient trajectory modelling, Yoon~\textit{et al.}~\cite{yoon2018invase} and Zhang~\textit{et al.}~\cite{zhang2020SMS_DKL} addressed two important aspects for predictive models in healthcare, namely variability in feature importance across samples and variability in optimal model selection throughout time, respectively. Yoon~\textit{et al.}~\cite{yoon2018invase} proposed INVASE with the objective of discovering flexible feature subsets instance-wise, using this solution for mortality prediction. INVASE is composed of three Fully Connected Networks~(FCNs) which are trained and optimized following an actor-critic reinforcement learning strategy, taking structured data as input. Despite not directly modelling temporal information, the authors proposed replacing each FCN with a Recurrent Neural Network~(RNN) to capture the temporal dynamics in data~\cite{yoon2018invase}. In~\cite{zhang2020SMS_DKL},  Zhang~\textit{et al.} proposed a system for automatically selecting the optimal predicting model for each time step, which was evaluated on the prediction of diseases, readmission and mortality. The model architecture consisted of a~RNN, a DeepSets layer that summarily consists of permutation invariant functions, a multilayer perceptron and a Bayesian linear regressor, and used \textit{T} optimizers, where \textit{T} corresponds to the number of time steps, to select the optimal model for each time step. In this work, the authors explored temporal data such as time series (\textit{e.g.} vital signs) and modelled time by having different prediction models for different time steps (stepwise model selection)~\cite{zhang2020SMS_DKL}.

Aiming at the early detection of acute critical illness such as sepsis, acute kidney injury and acute lung injury, Lauritsen~\textit{et al.}~\cite{lauritsen2020explainableAI} used a Temporal Convolutional Network~(TCN) based model with a Deep Taylor Decomposition~(DTD) explanation module that was adjusted for temporal explanations. This solution explored structured information and time series data as input, and addressed the temporal component through several strategies such as using only information from a given time window, binning events in intervals of 1 hour, and including temporal state pooling in the TCN part of the model~\cite{lauritsen2020explainableAI}. Si~\textit{et al.}~\cite{Si2020} used a three-level Hierarchical Attention Network~(HAN) where each level comprised an encoder (BiLSTM) and an attention layer. Patient embeddings were extracted from clinical notes and used as model input, using a list of 15 International Classification of Diseases~(ICD)-9 codes to define obesity related classification outcomes, and clinical notes were clustered within the patient timeline using a greedy algorithm with cluster interval time spans ranging from 30 minutes to 36 hours~\cite{Si2020}. In BEHRT, Li~\textit{et al.}~\cite{li2020behrt} used Bidirectional Encoder Representations from Transformers~(BERT) to represent EHR information from patient visits combined with a classification layer to predict the diseases in the next visit/next 6 to 12 months~\cite{li2020behrt}. This model used structured data from coding standards as input and integrated time through the inclusion of patient age in the input features~\cite{li2020behrt}. Rodrigues-Jr~\textit{et al.}~\cite{LIGDoctor2020} created LIG-Doctor as a solution for next visit diagnosis prediction, which uses two parallel Minimal GRU~(MGRU) layers working in opposite directions, acting as a bidirectional MGRU layer (similar to a BiLSTM), followed by two feed-forward layers. While the model is centered on structured data from coding standard as input, admission duration and time elapsed between admissions were also explored as input features to capture time~\cite{LIGDoctor2020}.

AdaptiveNet was created by H\"ugle~\textit{et al.}~\cite{AdaptiveNet2020} as a flexible architecture capable of supporting the existing multimodality in clinical data, and was used for disease progression modelling. The model consists of a set of modality-specific encoders, a LSTM that pools events and computes fixed-length encoded patient histories, and finally a linear layer. Although the authors only explored structured data in their work, the architecture can accomplish a \textit{straightforward integration} of multimodal information by encoding each modality using an arbitrary encoder such as a FCN, a RNN, a Convolutional Neural Network~(CNN) or other, enforcing a key constraint that all encoders have a final shared layer so that all modalities are projected into the same latent/encoded event space~\cite{AdaptiveNet2020}. Franz~\textit{et al.}~\cite{franz2020DeepObserver} proposed DeepObserver as a model for diagnosis prediction, consisting of a CNN followed by three FCN layers. However, the CNN was only selected for the first layer after evaluating the model with a FCN, a CNN and a RNN in the first layer, and observing that the CNN returned the best performances. DeepObserver combined structured information with time series data in its input, and employed a binning strategy to group numerical observations into 4 different bins considering the timing of each sample within the admission timeline.~\cite{franz2020DeepObserver}.

With recent development on language models, researchers have sought new approaches to more directly explore free text as a relevant data source, namely through the use of language models such as BERT to convert text into dense representations. Within this scope, Huang~\textit{et al.}~\cite{clinicalBERT2020} proposed clinicalBERT, a BERT model that was pre-trained on a clinical notes corpus and fine-tuned on the clinical task of predicting patient readmission. This work explored two types of clinical text sources, discharge summaries and early notes (first 2 and 3 days after admission) from patient admissions~\cite{clinicalBERT2020}. Franz~\textit{et al.}~\cite{franz2020DeepObserver} proposed a variation of clinicalBERT~\cite{clinicalBERT2020}, maintaining most of the model architecture but changing the downstream task from readmission to disease diagnosis prediction~\cite{franz2020DeepObserver}. Similarly to its parent model, clinicalBERT\_multi also uses free text reports as input, with the authors comparing the impact of using discharge summaries and early notes in model performance.

\section{Materials and Methods}
\label{sec:Methods}

This work used the MIMIC-III clinical database~\cite{Johnson2016} as it is publicly available and has been widely used throughout the research community to develop clinical decision support systems, enabling a more straightforward benchmarking with other existing solutions. Notably, this dataset has been used in the development of patient trajectory modelling solutions to predict clinical outcomes~\cite{LIGDoctor2020,clinicalBERT2020,franz2020DeepObserver}.

MIMIC-III is a large structured dataset containing large amounts of multimodal information collected from over 40 thousand patients in an intensive care unit scenario. Stored data encompasses clinical notes written by care providers, vital signs, laboratory measurements, diagnostic and procedure codes, patient demographics, medication information, among many other types of information. Furthermore, some of this information is standardised through the use of coding standards, for instance diagnostic and procedure codes come from the ninth revision of ICD, and medication information leverages the RxNorm standard lexicon.

The developed system only explored a feature subset from the wide variety of information modalities stored in MIMIC-III. The selected subset comprised three different types of features, including textual features, temporal features and coding features. The text type consists of clinical notes (discharge summaries) written by the medical staff. For temporal features, we selected the duration of the patient admissions as well as elapsed days between each admission. Finally, regarding coding features, both diagnosis and procedure ICD-9 codes were extracted, along with the list of medication codes associated with each patient visit.

Since selected data was used to develop trajectory modelling systems focused on distinct clinical outcomes, different data preprocessing procedures had to be used for each system. The resulting preprocessing pipeline is described in detail in Section~\ref{chap5:methods:datapreparation}.

\subsection{Data Preparation}
\label{chap5:methods:datapreparation}

The proposed solution is designed for the prediction of two fundamentally different clinical outcomes, requiring two different datasets. Since the diagnosis prediction task requires a dataset with higher specificity, a core preprocessing pipeline is firstly used to generate the readmission prediction dataset (based on the work from Huang~\textit{et al.}~\cite{clinicalBERT2020}), and then further processing steps are added to this pipeline for the creation of the diagnosis prediction dataset.

\subsubsection{Extracting Data From MIMIC-III}
\label{methods:datapreparation:mimic}

The datasets are composed of three major feature types: time-related, textual, and coding features. The pipeline begins with the creation of time-related features, computing elapsed days since the previous visit for each patient admission, as well as admission duration (in days). The readmission prediction task is focused on detecting more urgent patient readmissions, thus we set a threshold period of 30 days between a patient discharge and an unexpected readmission,~\textit{i.e.} samples are only labeled as positive for the readmission prediction task if less that 30 days have passed since the previous patient discharge. 

The next step regards the extraction and processing of coding information from MIMIC-III. Firstly, diagnoses ICD-9 codes are extracted
from the \textquotedblleft Diagnoses\_ICD\textquotedblright~table, ordered according to the \textquotedblleft Seq\_Num\textquotedblright~attribute, and saved as a list of ICD-9 codes for each patient admission. The same process is applied to the \textquotedblleft Procedures\_ICD\textquotedblright~table to extract procedure ICD-9 codes from patient admissions. Medication information is extracted from the \textquotedblleft Prescriptions\textquotedblright~table, ordered according to the start date attribute, and saved as a list of National Drug Codes~(NDCs) per admission.

The use of clinical text as an input feature is inspired by Huang~\textit{et al.}~\cite{clinicalBERT2020}, who used clinicalBERT (a BERT-based model) for representing clinical text in a clinical outcome prediction system. The authors provide two publicly available versions of the model\footnote{\url{https://github.com/kexinhuang12345/clinicalBERT}}: a pretrained version where BERT was pretrained using clinical notes from the MIMIC-III dataset, and a fine-tuned version where the pretrained model was fine-tuned for a readmission prediction task. Since we use clinicalBERT to convert clinical text to dense feature representations, a similar text processing strategy is followed where discharge notes are extracted from the \textquotedblleft Noteevents\textquotedblright~table, and the text from these notes is preprocessed using the following pipeline: carriage returns and line breaks are deleted, all text is changed to lowercase, special characters and symbols are cleaned and punctuation removed from abbreviated expressions such as \textquotedblleft dr.\textquotedblright~and \textquotedblleft m.d.\textquotedblright. Since clinicalBERT has an input size of 512, preprocessed clinical notes are segmented and saved in various text parcels as described in~\cite{clinicalBERT2020}, resulting in patient admissions with a variable number of clinical note chunks.

The final step at this stage consists in properly distributing and saving the resulting dataset. Since the objective of this work is to model patient trajectories, we consider that admissions from the same patient are not independent and as such should be stored together to avoid cross-talk between data splits (train, validation and test splits) during model development, contrasting to the approach used in~\cite{clinicalBERT2020,franz2020DeepObserver} where all admissions were considered independent and thus distributed through the data splits with no constraint. At this stage, the dataset is composed of 34~560 patients, from which 2~263 have been readmitted within the 30-days period at least once (positive samples for readmission prediction), whereas 32~297 patients have not (negative samples). Considering this statistic, patients are distributed using a 10-fold stratified strategy, resulting in a similar distribution of negative-positive patients per fold and ensuring that a patient does not appear in more than one fold. Next, each patient is expanded to the corresponding admissions, and admissions are shuffled resulting in the \textquotedblleft final\textquotedblright~readmission prediction dataset.

The diagnosis prediction dataset is obtained through further data processing and selection over the readmission prediction dataset. The readmission prediction dataset contains 28~173 subjects with a single admission and 6~387 subjects with multiple visits registered in the MIMIC-III database. The objective of predicting diagnoses codes for a future patient admission implies the existence of a subsequent visit with annotated diagnoses codes to serve as gold standard label. Therefore, only patients with multiple visits are admissible for this dataset, greatly reducing dataset size. Moreover, as the most recent admission for each patient does not have a list of future diagnoses codes, these admissions cannot be used for model development and thus have to be removed from the dataset. Finally, label generation for the diagnosis prediction task must follow a different procedure since this consists in a multi-label multi-class prediction task. The process developed for the generation of expected labels for this task is described later in this Section as it depends on additional processing steps. Table~\ref{tab:chap5:hadm_distribution} summarises the distribution of each dataset, reporting the number of patient admissions present in each of the 10 generated folds. It should be noted that each fold effectively contains more samples, as each admission has its discharge summary segmented in a variable number of chunks, resulting in an actual number of samples higher than the values reported in Table~\ref{tab:chap5:hadm_distribution}.  

\subsubsection{Preprocessing Data For The DL Models}
\label{chap5:methods:datapreprocess}

\paragraph{Temporal Features} \hspace{1pt}
With the relevant information extracted from the MIMIC-III database, it is necessary to follow through with additional preprocessing mechanisms to prepare the data for the DL model. Temporal features undergo light preprocessing, requiring only a feature rescaling to have their values ranging within the real interval of $[0,~1]$.

\paragraph{Text Features} \hspace{1pt}
The processing mechanism for clinical text was influenced by a key decision on model architecture. Despite initially intending to use the clinicalBERT model in a fine-tuning based approach (as in~\cite{clinicalBERT2020,franz2020DeepObserver}), where additional components would be added to the model and trained, this was not possible due to computational resource restrictions. Instead, clinicalBERT is used in a feature-based approach to convert text from clinical notes into dense representations ready to be used in other DL models. 
Since in practical terms this corresponds to having a frozen model generate fixed representations for a textual feature during every epoch, which naturally still implies a major time overhead, we precompute the clinicalBERT representations of every clinical text chunk and save them in disk. By having the textual feature representations ready for access in memory, it is possible to simply load these representations whenever necessary and use them as model input, significantly accelerating the process of model developing.

\paragraph{Coding Features} \hspace{1pt}
Coding features require a more thorough preprocessing pipeline owing to 1) inherent characteristics associated with this type of data, 2) decisions on model architecture and 3) the fact that they constitute the core resource in label generation for the diagnostic prediction task. ICD-9 contains over 22~200 different codes, which can be divided in diagnostic (17~552 ICD-9-CM codes) and procedure (4~651 ICD-9-PCS codes) categories. However, only a subset of these are present in MIMIC-III, which contains approximately 7~000 ICD-9 diagnostic codes. Since using a large search space in prediction tasks can be prohibitive for the process of model training, negatively impacting on model performance, a possible approach is to convert original codes to a more simplified coding standard at the cost of reducing information granularity. A common alternative when dealing with ICD-9 codes is to map ICD using Clinical Classifications Software~(CCS) resources from the Healthcare Cost and Utilization Project\footnote{\url{https://www.hcup-us.ahrq.gov/toolssoftware/ccs/ccs.jsp}}. CCS clusters ICD into a considerably smaller scope of clinical codes, narrowing down diagnoses to approximately 250 codes and procedures to nearly 230 codes. Following similar works in this area~\cite{LIGDoctor2020, franz2020DeepObserver}, all ICD-9 codes (diagnoses and procedures) are mapped to CCS.

Since reducing possible codes by two orders of magnitude ($10^4$ to $10^2$) can lead to the loss of too much granularity in information, an additional processing step was implemented to create a condensed version of ICD. This step firstly differentiates diagnosis and procedure codes by prepending a \textquotedblleft D\_\textquotedblright~or \textquotedblleft P\_\textquotedblright~string, and then extracts the parent part of the code according to the type of code being processed: the base structure of diagnoses codes can follow three different root forms --- 3 digits (xxx) for normal diagnosis codes; E followed by 3 digits (Exxx) for a branch of supplementary codes; and V followed by 2 digits (Vxx) for another branch of supplementary codes ---, whereas procedure codes have a simpler structure which always starts with two digits at the parent level. By reducing ICD-9 codes to their parent level, it is possible to condense ICD-9 into 1~234 diagnosis codes and 100 procedure codes (lowering one order of magnitude). This results in a smaller search space than that of complete ICD-9, which still retains more detailed information than CCS codes.

Medication information (NDCs) extracted from MIMIC follows a similar procedure as ICD codes to reduce vocabulary dimension. NDCs are very specific as they hold detailed information such as the strength and packaging of the medication being coded, resulting in a vocabulary with over 250~000 codes. Nonetheless, each NDC code is associated with a Concept Unique Identifier~(CUI) through Unified Medical Language System~(UMLS). This relation is leveraged to extract the corresponding CUIs, leading to a reduced vocabulary size slightly below 20~000 codes.

Due to the natural variability associated with patients' health status, each admission can list a varying number of diagnoses, procedures and medications. Padded tensors are created to handle variability, storing this admission information into vectorial structures with a size equal to the maximum number of different codes (for each category) present in a patient admission.

Furthermore, the decision to use SapBERT~\cite{sapbert2021} to represent coding information, which is explained in more detail in Section~\ref{chap5:sec:modelarchitecture}, demands an additional processing step on coding information. SapBERT is a publicly available BERT model that was pretrained on UMLS, obtaining state-of-the-art results in medical entity linking problems. This model takes very small sentences as input (BERT tokenized sentences with a maximum length of 25 tokens) but its output has the same dimensions as base BERT models (768). To obtain SapBERT representations, text descriptions are firstly extracted from the UMLS for ICD-9 and CUI, and directly from CCS resources for CCS codes. These descriptions are then fed to SapBERT and the resulting embedding representations are saved to be later used as an embedding layer in the proposed model architecture.

The final part of the pipeline for coding features concerns label generation for the diagnosis prediction task. Since patients can suffer from multiple diseases (comorbidities) and these diseases can co-occur, the problem of modelling future diagnoses is addressed from a multi-label multi-class perspective, enabling the prediction of multiple diagnosis codes. Thus, to generate expected labels, each admission has its list of future diagnoses codes converted into a multi-hot vector with dimension equal to the number of possible diagnoses codes to predict (\textit{e.g.} 1~234 for ICD-9 diagnoses codes), where codes present in the next admission are set to 1 and non existing codes are left at 0.

\paragraph{Sequence Modelling}

The objective of this work was to explore the patient medical history,~\textit{i.e.} the patient trajectory, therefore a recurrent model architecture was implemented that considers temporal sequences of patient visits instead of isolated admissions in the prediction process. For the readmission task, the previous readmission dataset is firstly filtered to remove patients with only a single admission, since the concept of trajectory presupposes the existence of more than one visit, followed by a sorting by admit time (per patient). Then, sequences of admissions, containing either 3 or 6 admissions, are packed into a vectorial representation using a sliding window mechanism that iterates through the admission history of each patient. Finally, dense representations are precomputed for the patient admissions using a previously developed model from readmission prediction using individual admissions. The creation of admission sequences for diagnoses prediction follows a similar procedure differing only in: the removal of single-visit patients which has already been performed \textit{a priori}, and the precomputation of dense representations which is performed using a previously developed diagnoses prediction model.

\subsection{DL Model Development}
\label{chap5:sec:modelarchitecture}

\subsubsection{Model Architecture}

The proposed solution involved the development of two different model architectures. The first one was designed with the objective of predicting patient readmission and diagnoses using isolated admission views, whereas the second one explores sequences of visits to model the patient trajectories whilst leveraging from the first architecture to convert admissions into dense representations. Herein we focus on the former, providing more details on its implementation.

\begin{landscape}  
    \begin{figure}[!h]
        \centering
        \includegraphics[width=\linewidth]{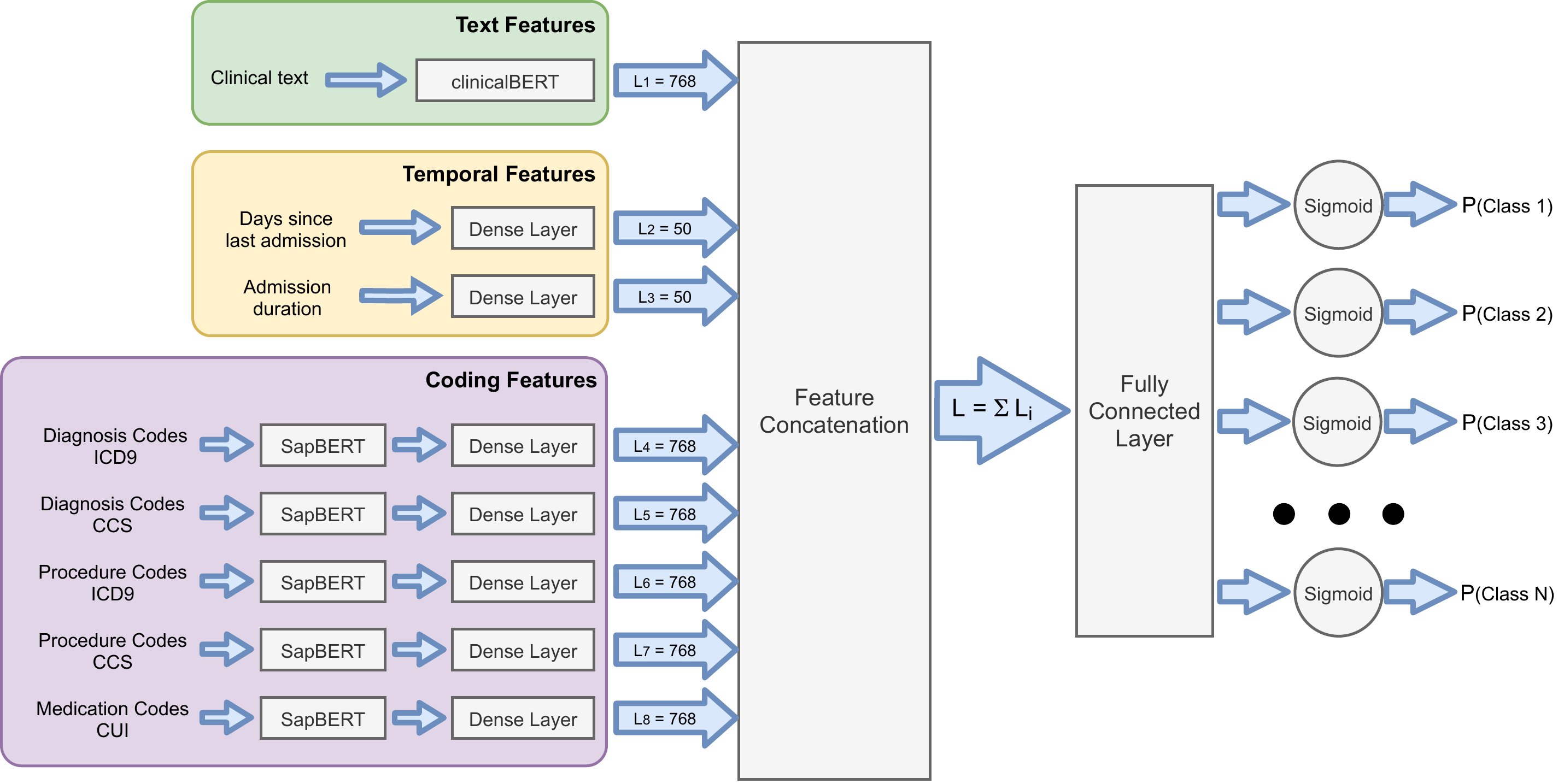}
        \caption[Schematic diagram of the first model architecture used for the readmission prediction and diagnosis code prediction downstream tasks.]{Schematic diagram of the model architecture used for the readmission prediction and diagnosis code prediction downstream tasks. The dimension of each feature representation is presented next to their respective layer. Since the model is flexible and can be trained using different feature subsets, the output from the feature concatenation layer varies in dimension, corresponding to the summation of the dimensions of used features. Finally, \textit{N} sigmoid activations are used to calculate class probabilities, where \textit{N} corresponds to the number of classes to predict. Since in readmission prediction only one class is predicted, the final part of the diagram can be simplified into a single sigmoid activation for that specific case.}
        \label{fig:chap5:modelarchitecture}
    \end{figure}
\end{landscape}

The readmission and diagnoses prediction model, which is presented in Figure~\ref{fig:chap5:modelarchitecture} as a schematic overview, integrates aspects from other existing works in the area, namely clinicalBERT~\cite{clinicalBERT2020}, clinicalBERT\_multi~\cite{franz2020DeepObserver} and AdaptiveNet~\cite{AdaptiveNet2020}. More specifically, the process of using contextual embeddings to model clinical notes in downstream tasks such as readmission and diagnoses prediction is based on~\cite{clinicalBERT2020,franz2020DeepObserver}, whilst the concept of having a shared latent space for different input features is based on~\cite{AdaptiveNet2020}. Additionally, the model is implemented with the objective of providing flexibility in feature selection, enabling model development with any of the features presented in Figure~\ref{fig:chap5:modelarchitecture} or with any feature combination. This selection directly impacts on the final dense representation of the patient admission being modelled (output of the Feature Concatenation layer in Figure~\ref{fig:chap5:modelarchitecture}).

Beginning with the text features component, clinicalBERT has shown potential in modelling clinical text for early readmission prediction~\cite{clinicalBERT2020} and for the detection of possible diagnosis codes associated with the patient admission~\cite{franz2020DeepObserver}. Since the objective was to explore the combination of information from clinical notes with other data modalities, we integrate clinicalBERT in the model architecture to convert clinical text into dense representations that can be leveraged by other parts of the proposed model. To avoid possible data leakage into the test partitions of the dataset, the pretrained version of the clinicalBERT model are used instead of its fine-tuned counterpart. 

Despite the initial idea to use clinicalBERT in a fine-tuning approach, both resource and time restrictions led to the decision of using it in a feature-based approach instead, with all of its layers being frozen. Furthermore, since this configuration of the clinicalBERT model provides non-mutable dense representations, it is possible to precompute the dense representations for each clinical note segment, thus avoiding the computational overhead of forwarding clinical note segments through clinicalBERT every epoch, and accelerating the training process of the whole model presented in Figure~\ref{fig:chap5:modelarchitecture}.

Temporal features involve a simple part of the model, being fed to simple dense layers with layer normalisation, Mish activation and dropout. The part of the model responsible for coding features is more complex, with coding features being forwarded through an embedding layer with layer normalisation and dropout, followed by a dense layer with layer normalisation, Mish activation function and dropout. The coding features component integrates the concept of shared representation spaces, using the same contextualized embeddings model (SapBERT) to represent the different types of codes (ICD, CCS and CUIs from UMLS). The decision to use SapBERT for the conversion of coding features into dense representations is based on the idea of projecting diagnoses codes into a representation space where semantically similar codes are closer together as it might help the model understanding possible relations between different diagnoses codes, which can be beneficial for the diagnoses prediction task.

Since text descriptions for ICD, CCS and CUIs represent a small portion of SapBERT's vocabulary, a similar procedure is followed as in clinicalBERT with the dense representations for all codes being precomputed and saved in disk. The resulting embeddings can then be loaded in smaller embedding layers in the proposed model architecture, serving as look-up tables for the input coding features which map codes to the corresponding embeddings, and reducing the overhead in computation time. Moreover, as the proposed model provides flexibility in feature selection, this approach enables the loading of only the required embeddings,~\textit{i.e.} embedding layers for the selected coding features, reducing unnecessary memory footprint.

The SapBERT embedding layers presented in Figure~\ref{fig:chap5:modelarchitecture} receive input tensors with groups of ICD, CCS and CUI codes, thus the output of these layers consists of groups of dense representations. Tensor length differs for each type of coding feature, which limits the use of a shared layer to condense all coding features into the same output dimension. Therefore, separate dense layers are used after each embedding layer acting as bottlenecks that project information into a tensor with dimension 768. All dense layers have layer normalisation, Mish activation and dropout. 

Dense representations from all selected features are concatenated in the next layer, hence the size of the final fused representation corresponds to the sum of the dimensions of selected features representations. A final FCN converts admission representations into an output with dimension equal to the number of classes to predict, which varies according to the prediction problem being modelled: readmission prediction only requires one class (binary classification), whereas diagnoses prediction requires a number of classes equal to the number of existing diagnostic codes (1~234 for ICD-9 codes and 255 for CCS codes).

The concluding piece of this model architecture concerns the selection of the final activation and loss function for the two different problems of readmission and diagnoses prediction. The Sigmoid activation function presented in Equation~\ref{eq:chap5:sigmoid}, which is typically used in binary classification applications, converts raw input values into values within the $[0,~1]$ range. Output values from this function can then be interpreted as the predicted probability of a given class being positive.

\begin{equation}
    \label{eq:chap5:sigmoid}
    \textsf{Sigmoid}(x)=\frac{1}{1+e^{-x}}
\end{equation}

The Sigmoid activation is usually paired with the binary cross entropy loss function. Binary cross entropy is presented in Equation~\ref{eq:chap5:binarycrossentropy}, where $N$ is the number of samples, $y_i$ corresponds to the true class of sample $i$ (being either 0 or 1), $p_i$ corresponds to the probability of being class 1, and 1-$p_i$ corresponds to the probability of being class 0.

\begin{equation}
    \label{eq:chap5:binarycrossentropy}
    \textsf{Binary Cross Entropy}=\frac{1}{N} \sum_{i=1}^N -(y_i \log(p_i) + (1-y_i)\log(1-p_i))
\end{equation}

However, when dealing with the modelling of multi-class problems, a different activation function is usually used which is the Softmax function (Equation~\ref{eq:chap5:softmax}). Softmax receives a vector of raw values with length $N$ (number of classes being modelled) as input, and converts each value into a class probability. However, when computing each class probability, Softmax normalises it by the sum of the exponentials of all $N$ classes, resulting in a list of predicted probabilities whose summation equals 1, and where the highest probability defines the predicted class.

\begin{equation}
    \label{eq:chap5:softmax}
    \textsf{Softmax}(x_i)=\frac{e^{x_i}}{\sum_{j=1}^N e^{x_j}},\; \textsf{for}\; j=1,...,N
\end{equation}

Softmax activation is typically used with the cross entropy loss function, presented in Equation~\ref{eq:chap5:crossentropy}, where $M$ is the number of samples, $N$ is the number of classes, $y_{ij}$ identifies (with 0 or 1) if class $j$ is the correct class for sample $i$, and $p_{ij}$ is the probability of sample $i$ belonging to class $j$.

\begin{equation}
    \label{eq:chap5:crossentropy}
    \textsf{Cross Entropy}=-\frac{1}{M} \sum_{i=1}^M \sum_{j=1}^N y_{ij}\log(p_{ij})
\end{equation}

Due to its suitability for binary classification problems, the Sigmoid activation is used with binary cross entropy for model development in readmission prediction. Regarding diagnoses prediction, despite being adequate for single-label multi-class prediction problems, Softmax activation is not suitable for multi-label multi-class prediction problems since it generates mutually exclusive outputs,~\textit{i.e.} it always predicts a single class as true (highest probability) at the cost of diminishing the probabilities of all remaining classes (which can also be true).

An alternative approach for the modelling of multi-label multi-class problems is to use the Sigmoid function with binary cross entropy, addressing this task as $N$ binary classification problems. Since the Sigmoid activation provides non-exclusive outputs, predicted probabilities for each class are independent and more than one class can be labeled as true. This combination is particularly useful for the diagnoses prediction problem as it can model different scenarios in the patient health status such as: the prediction of a single disease, the prediction of multiple diseases in patients with comorbidities, and also the prediction of no diseases. Therefore, Sigmoid activation and binary cross entropy are also used for model development in diagnoses prediction.

The resulting model architecture (Figure~\ref{fig:chap5:modelarchitecture}) converts input features from a patient admission into an admission representation (that we define as $h_{admission}$), and computes class predictions based on $h_{admission}$. However, since clinical notes have varying length and need to be partitioned in smaller text segments, the selection of clinical text as an input feature leads to a special scenario where a patient admission can be represented by several admission representations (each with a clinical note chunk) and not by a unique $h_{admission}$, hence generating class predictions for each sub representation. This group of predictions is condensed into a single prediction using Equation~\ref{eq:chap5:admissionaggregation} as proposed by Huang~\textit{et al.}~\cite{clinicalBERT2020}:

\begin{equation}
    \label{eq:chap5:admissionaggregation}
    P(Class\;N \mid h_{admission}) = \frac{p^k_{max} + p^k_{mean} * \lambda}{1 + \lambda},\;\textsf{where}\;\lambda = \frac{k}{c}
\end{equation}

Equation~\ref{eq:chap5:admissionaggregation} aggregates predicted probabilities for a patient admission containing $k$ clinical note segments, using the maximum probability $p^k_{max}$ among the $k$ predictions as well as the mean probability $p^k_{mean}$. As longer clinical notes (higher $k$) have a higher chance of containing a noisy maximum probability that can skew the final prediction, a scaling factor $\lambda$ is used to weight up the contribution of the mean probability and balance the numerator component, where $c$ is a constant that accounts for patients with a larger number of clinical text segments ($c$ was set to 2 as in~\cite{clinicalBERT2020}). The denominator of Equation~\ref{eq:chap5:admissionaggregation} is responsible for normalising the final prediction into a probability value within [0,~1].


\begin{landscape} 
    \begin{figure}[!h]
        \centering
        \includegraphics[width=\linewidth]{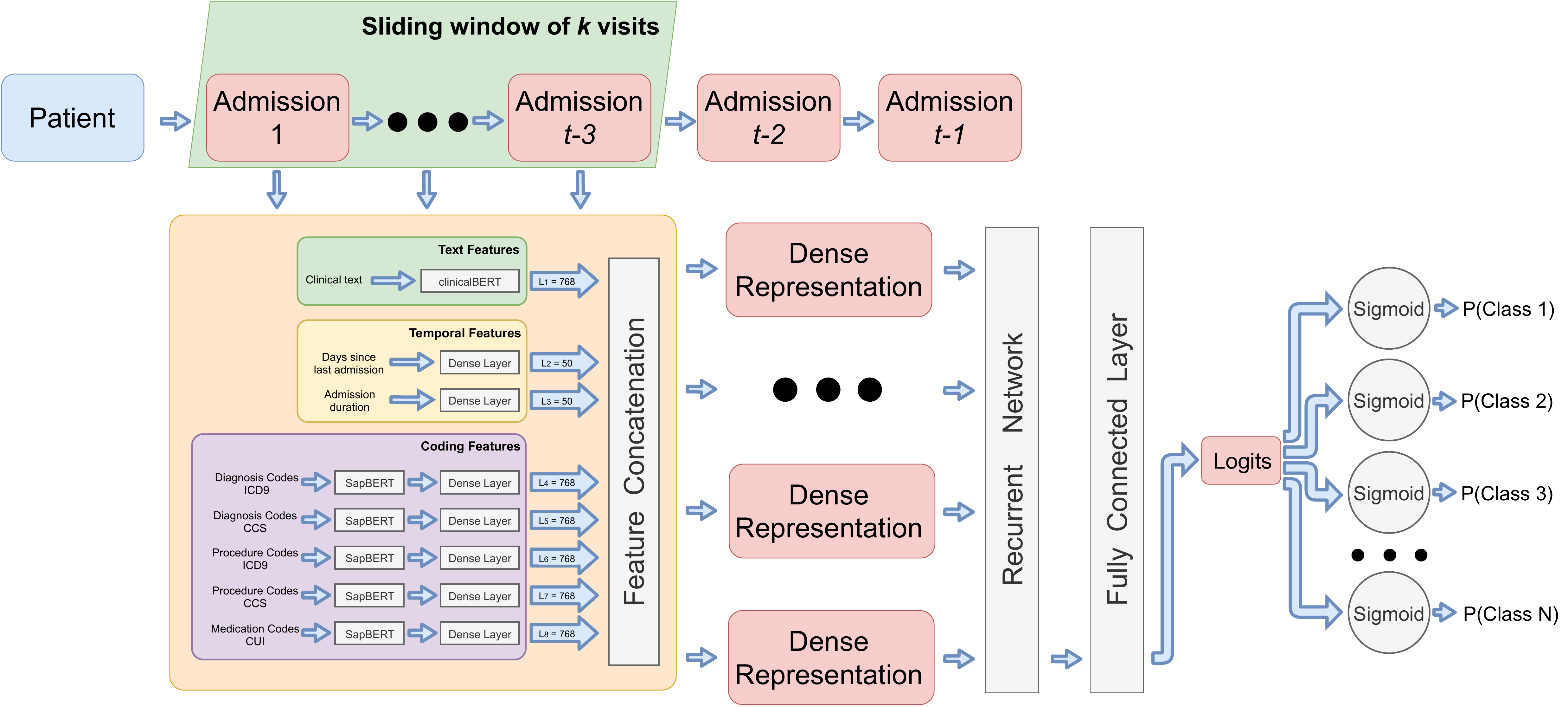}
        \caption[Schematic diagram of the sequence-based trajectory modelling architecture.]{Schematic diagram of the sequence-based trajectory modelling architecture. This architecture builds upon the previous readmission/diagnosis code prediction models, using them to generate dense representations (extracted from the \textquotedblleft Feature Concatenation\textquotedblright~layer) for each patient admission. Since there exist patients with a very long admission history, the model considers sliding windows with \textit{k} admissions for each patient, scanning from the oldest to the most recent admission. The resulting dense representations are forwarded through a recurrent model and a fully connected layer. Finally, \textit{N} sigmoid activations are used to calculate class probabilities, where \textit{N} corresponds to the number of classes to predict. Since in readmission prediction only one class is predicted, the final part of the diagram can be simplified into a single sigmoid activation for that specific case.}
        \label{fig:chap5:trajectorymodelarchitecture}
    \end{figure}   
\end{landscape}

The first model architecture was designed to convert information from patient admissions into dense representations and generate class predictions solely based on individual patient admissions, using isolated points in time for each prediction and ignoring the patient medical history altogether. Since one of the main objectives was to explore the patient trajectory, a second model architecture was designed that leverages the previous one and explores the full patient history. A schematic view of the trajectory modelling architecture is presented in Figure~\ref{fig:chap5:trajectorymodelarchitecture}.

The proposed architecture uses a recurrent network to consider temporal sequences of patient visits in the prediction process, instead of individual admissions. Due to the existing variability in terms of number of admissions per patient in the MIMIC-III database, where there exist patients with 40 registered admissions in their medical history, a sliding window mechanism is integrated that iterates through the patient medical history from the oldest to the most recent usable admission ($t-1$) and feeds the model with sequences of $k$ admissions. When patients have less than $k$ admissions in history, the sliding window is padded with dummy admissions to ensure its fixed size. The last admission (time $t$) cannot not be used for model training as it does not have information from a future admission, hence being only used to generate the correct labels for predictions at admission $t-1$.

Each admission in the sliding window is forwarded through the previous model architecture (Figure~\ref{fig:chap5:modelarchitecture}), and the corresponding dense representations are extracted from the feature concatenation layer. It should be noted that the selection of the model responsible for converting each admission into $h_{admission}$ is performed on a task specific basis,~\textit{i.e.} instances of the model from Figure~\ref{fig:chap5:modelarchitecture} that were previously trained on readmission prediction and diagnoses prediction are leveraged in Figure~\ref{fig:chap5:trajectorymodelarchitecture} to generate the corresponding admission representations.

Generated admission representations are then used as input in the recurrent neural network. Two different types of networks were implemented for this stage, GRU and LSTM, each supporting bidirectionality. The output of the recurrent network is passed to a fully connected layer that converts representations into tensors of dimension $N$ (number of possible classes). Finally, Sigmoid activation and binary cross entropy are used both for readmission prediction and diagnoses prediction.

\subsubsection{Model Training}

The first model architecture was trained for the two tasks of readmission and diagnoses prediction using the same hyperparameters settings, which are presented in Table~\ref{tab:chap5:hyperparameters_individualmodel}, differing only in the use of data subsampling. Data subsampling is used in readmission prediction to enable a more direct comparison with clinicalBERT, which was fine-tuned for readmission prediction using a balanced dataset where negative samples were downsampled to match the number of positive samples in MIMIC-III. However, while Huang~\textit{et al.}~\cite{clinicalBERT2020} performed class balancing using a small set of negative admissions that was defined \textit{a priori}, we use a different approach where in each fold of cross-fold validation, negative samples are randomly sampled from a large pool of negative admissions, matching the number of positive samples in the dataset. Concerning diagnoses prediction, although Franz~\textit{et al.}~\cite{franz2020DeepObserver} used an iterative sampling scheme to ensure that each data split (train-val-test) shared the same label distribution, that approach would compromise the key premise of not letting information from the same patient spread throughout different data splits, thus no sampling scheme is used.

\begingroup
\renewcommand{\arraystretch}{1.2}
\begin{table}[!t]
\centering
\caption[List of~\textquotedblleft optimal\textquotedblright~hyperparameters used for training the first model architecture.]{List of~\textquotedblleft optimal\textquotedblright~hyperparameters used for training the first model architecture. RP: readmission prediction, DP: diagnoses prediction.}
\label{tab:chap5:hyperparameters_individualmodel}
\begin{tabular}{ccc}
\toprule
Description & & Value \\
\cmidrule{1-1}\cmidrule{3-3}

Dimension of clinicalBERT embeddings & & 768\\
Dimension of SapBERT embeddings & & 768\\
Hidden layer size (temporal features) & & 50 \\
Dropout rate & & 0.1\\
Activation function & & Mish\\
Subsampling & & RP: True | DP: False\\
Train batch size & & 64\\
Number of training epochs & & 100\\
Learning rate & & 0.001\\
Early stop & & True\\
Number of patience steps & & 200\\
Optimizer & & Ranger\\

\bottomrule
\end{tabular}
\end{table}
\endgroup

\begingroup
\renewcommand{\arraystretch}{1.2}
\begin{table}[!h]
\centering
\caption[List of experimental hyperparameters for training the second model architecture.]{List of experimental hyperparameters used for training the second model architecture.}
\label{tab:chap5:hyperparameters_trajectorymodel}
\begin{tabular}{ccc}
\toprule
Description & & Value \\
\cmidrule{1-1}\cmidrule{3-3}
Type of recurrent network & & \{GRU, LSTM\}\\
Hidden layer size (recurrent network) & & \{255,~818\}\\
Number of layers (recurrent network) & & \{1,~2,~3\}\\
Bidirectionality & & \{True, False\} \\
Dropout rate & & 0.2\\
Admissions in sliding window & & \{3,~6\}\\
Train batch size & & 64\\
Number of training epochs & & 200\\
Learning rate & & 0.001\\
Early stop & & True\\
Number of patience steps & & 200\\
Optimizer & & Ranger\\

\bottomrule
\end{tabular}
\end{table}
\endgroup

Table~\ref{tab:chap5:hyperparameters_trajectorymodel} presents a list of experimental hyperparameters used in the recurrent model architecture. 
All models herein presented were trained using early-stopping to reduce model overfitting. Additionally, Weights \& Biases was integrated in this work to facilitate model prototyping and experiment tracking~\cite{wandb}.

\section{Results}
\label{sec:results}

\subsection{Initial Experiments}
\label{results:initial}

Initial experiments were performed with the first model architecture using a fine-tuned version of clinicalBERT and a 5-fold version of the original dataset from~\cite{clinicalBERT2020}. Model performance was evaluated using the following list of metrics: precision, recall, $F_1$-score, AUC, AU-PR and RP80, where AUC corresponds to the area under the receiver operating characteristic curve, AU-PR corresponds to the area under the precision-recall curve, and finally RP80 corresponds to recall at a precision of 80\%. RP80 was used as a more clinically-relevant evaluation metric since it assessed model performance from the perspective of maximizing the retrieval of relevant predictions whilst minimizing false positive rate. Precision, recall and $F_1$-score were computed using a threshold of 0.5 for the classification of negative/positive classes, whereas AUC, AU-PR and RP80 were calculated using variable thresholds. Model development for the first experiment was performed using a 5-fold cross-validation strategy where three folds were used for training, one fold was used for validation (to train the model using early-stopping) and one fold was used to test the resulting model. Obtained results are presented at the top half of Table~\ref{tab:chap5:results_individual_readmission_experiments1}.

The dataset used in the first group of experiments was based on an~\textit{a priori} sampling of negative samples, significantly reducing the pool of possible negative samples from MIMIC-III. Therefore, for the second group of experiments, we created a new dataset where positive and negative admissions were extracted from MIMIC-III and distributed into 10 folds in a stratified manner. Stratified sample distribution was performed at the admission level, resulting in a more even class distribution across folds as shown in Table~\ref{tab:chap5:hadm_distribution_hadmsplit}, differently from Table~\ref{tab:chap5:hadm_distribution} where the distribution was performed at the patient level.

\begin{landscape}

\begingroup
\renewcommand{\arraystretch}{1.0}
\begin{table}[!th]
\centering
\caption[Experimental readmission prediction results for the individual admission model architecture, using fine-tuned clinicalBERT with a 5-fold version of the dataset from clinicalBERT.]{Experimental readmission prediction results for the individual admission model architecture, using fine-tuned clinicalBERT with a 5-fold version of the dataset from~\cite{clinicalBERT2020}, and pretrained clinicalBERT with the new 10-fold dataset. Best results in bold.}
\label{tab:chap5:results_individual_readmission_experiments1}
\begin{tabular}{clc cccccc@{}}
\toprule
Resources & Features & & AUC & AU-PR & $F_1$-score & Precision & Recall & RP80\\
\cmidrule{1-2}\cmidrule{4-9}
\multirow{5}{*}{\begin{tabular}[c]{@{}c@{}}Finetuned\\clinicalBERT\\ + \\ 5-fold original \\ dataset\end{tabular}} & Text & & 0.829 & 0.804 & 0.719 & 0.686 & 0.755 & 0.614 \\
 & Duration & & 0.611 & 0.581 & 0.504 & 0.531 & 0.482 & 0.025 \\
 & Days to prior admission & & 0.810 & 0.712 & 0.836 & 0.718 & \textbf{0.999} & 0.033 \\
 & Days to prior admission + Duration & & 0.864 & 0.780 & 0.836 & 0.718 & \textbf{0.999} & 0.238 \\	
 & Days to prior admission + Duration + Text & & \textbf{0.914} & \textbf{0.890} & \textbf{0.849} & \textbf{0.761} & 0.959 & \textbf{0.904}\\
\cmidrule{1-2}\cmidrule{4-9}
\multirow{7}{*}{\begin{tabular}[c]{@{}c@{}}Pretrained\\clinicalBERT + \\ 10-fold new \\ dataset\end{tabular}} & Text & & 0.822 & 0.403 & 0.713 & 0.683 & 0.747 & 0.001 \\		
& Duration & & 0.549 & 0.468 & 0.367 & 0.316 & 0.481 & 0.011 \\
& Days to prior admission & & 0.798 & 0.731 & 0.844 & 0.730 & 0.999 & 0.082 \\		
& Days to prior admission + Text & & 0.885 & 0.504 & \textbf{0.845} & 0.756 & \textbf{0.957} & 0.001 \\	
& Days to prior admission + Duration & & 0.788 & \textbf{0.761} & 0.805 & 0.730 & 0.898 & \textbf{0.358} \\		
& Days to prior admission + Duration + Text  & & \textbf{0.889} & 0.502 & \textbf{0.845} & \textbf{0.757} & 0.955 & 0.002 \\
\bottomrule
\end{tabular}
\end{table}
\endgroup
\end{landscape}

Model development for the second group of experiments was performed using 10-fold cross-validation strategy where eight folds were used for training, one fold was used for validation (to train the model using early-stopping) and one fold was used to test the resulting model. A subsampling procedure was used in each cross-validation iteration to randomly sample negative admissions from the data partitions, so that positive and negative classes had the same support. Since a different dataset was being used, a pretrained version of clinicalBERT was explored in this experiments instead of its fine-tuned counterpart. Obtained results are presented at the bottom half of Table~\ref{tab:chap5:results_individual_readmission_experiments1}.

During the initial experimentation phase, several tests were also performed with the proposed model architecture to assess the impact of varying the dimension of temporal features (\textit{i.e.} the size of the dense layer for temporal features) in model performance. A progressive subset of dimensions was tested for both temporal features: \{1, 2, 5, 10, 20, 50, 100, 200, 500\}. Experimental tests demonstrated that using reduced dimensions had a negative impact on model performance, whereas too large dimensions did not necessarily bring performance improvements. Therefore, a final layer dimension of 50 was selected for temporal features, as this balanced predictive performance with computational overhead. All results presented in Section~\ref{sec:results} which involve the use of temporal features were obtained using a layer size of 50 in the temporal dense layers.

\subsection{Readmission Prediction}
\label{chap5:results:readmission}

Previous experiments provided a base knowledge on how the proposed model architecture worked as well as on the potential impact of each input feature in the predictive power of the model. Nevertheless, as previously mentioned the datasets used in Section~\ref{results:initial} were created based on the assumption of each patient admission being independent, hence enabling the distribution of admissions from the same patient among different dataset folds, and creating a potential source of information leak in the model from train to test time.

\begin{landscape}
    \begingroup
\renewcommand{\arraystretch}{1.0}
\begin{table}[!ht]
\centering
\caption[Final readmission prediction results for the individual admission model architecture.]{Final readmission prediction results for the individual admission model architecture. Results reported for 10-fold cross-validation. ClinicalBERT results presented for comparison. Best results highlighted in bold.}
\label{tab:chap5:results_individual_readmission}
\begin{tabular}{lc cccccc}
\toprule
Features & & AUC & AU-PR & $F_1$-score & Precision & Recall & RP80\\
\cmidrule{1-1}\cmidrule{3-8}
Text  & & 0.795 & 0.363 & 0.717 & 0.680 & 0.759 & 0.001 \\
Duration & & 0.567 & 0.521 & 0.517 & 0.553 & 0.492 & 0.050 \\
Days to prior admission & & 0.797 & 0.739 & 0.822 & 0.737 & 0.931 & 0.373 \\
Text + Duration & & 0.794 & 0.360 & 0.719 & 0.680 & 0.762 & 0.001 \\
Days to prior admission + Text & & \textbf{0.871} & 0.468 & 0.820 & \textbf{0.766} & 0.881 & 0.001 \\
Days to prior admission + Duration & & 0.791 & \textbf{0.750} & \textbf{0.828} & 0.737 & \textbf{0.945} & \textbf{0.481} \\
Days to prior admission + Duration + Text & & 0.869 & 0.473 & 0.821 & \textbf{0.766} & 0.885 & 0.001 \\
Days to prior admission + Duration + Text + Diag ICD9 & & 0.773 & 0.440 & 0.739 & 0.733 & 0.754 & 0.001 \\
\cmidrule{1-1}\cmidrule{3-8}
ClinicalBERT~\cite{clinicalBERT2020} --- Text & & 0.714 & 0.701 & --- & --- & --- & 0.242 \\ 
\bottomrule
\end{tabular}
\end{table}
\endgroup    
\end{landscape}

However, we consider the previous view fundamentally flawed as it would consider each admission independent from the past medical state of the patient. Instead, we consider that admissions in a patient trajectory are not independent as they originate from the same source (the patient). 
Therefore, a new dataset was created with samples being distributed across data folds at a patient level, ensuring that admissions from each patient would not leak into different dataset partitions.

The resulting dataset was used for model development using the same 10-fold cross-validation strategy as before. Similarly to the experiments presented at the bottom half of Table~\ref{tab:chap5:results_individual_readmission_experiments1}, pretrained clinicalBERT was used, and a subsampling procedure was applied in each cross-validation iteration to randomly sample negative admissions and level class distribution. Obtained results are presented in Table~\ref{tab:chap5:results_individual_readmission}.

\subsection{Diagnoses Prediction}

Readmission prediction and diagnoses prediction are very distinct problems, thus shifting to a problem of modelling diagnoses predictions for future visits had implications both on the model and on the dataset. Firstly, diagnoses prediction consisted in a multi-label multi-class classification problem, naturally having higher complexity than the binary classification problem of readmission prediction. Nonetheless, owing to its multi-label component, instead of treating it as a normal multi-class problem we had to evaluate it as multiple binary classification problems. Doing so not only enabled the presence of multiple different classes in the same prediction for a patient admission, but also simplified the process of adapting the model to the new prediction task since both the activation and loss function were maintained. Regarding the second aspect, a new dataset had to be prepared for diagnoses prediction (using the procedure that was previously described in Section~\ref{methods:datapreparation:mimic}) since this task was inherently reliant on the concept of trajectory,~\textit{i.e.} patients had to have multiple admissions to be eligible.

Similarly to the previous prediction task, the dataset was used for model development using the same 10-fold cross-validation strategy, and pretrained clinicalBERT was used to represent clinical text. Despite the low cardinality of some diagnoses classes present in the output labels, no subsampling procedures were applied in this dataset. Furthermore, due to the distinct nature of this prediction problem, different metrics had to be used to assess model performance, namely: AU-PR, $F_1$-score, RP80, Recall@10, Recall@20 and Recall@30. The Recall@\textit{k} metric computes how many of the relevant diagnoses codes were retrieved in the list of top \textit{k} predicted diagnoses, as presented in Equation~\ref{eq:chap5:recallatk} where $\|y_{true}\|$ corresponds to the number of positive diagnoses codes in the ground-truth label.

\begin{equation}
    \label{eq:chap5:recallatk}
    Recall@k=\frac{\text{Correctly recommended codes in top \textit{k} predictions}}{\|y_{true}\|}
\end{equation}

Adjusting the value of~\textit{k} in Recall@\textit{k} leads to a trade-off where increasing~\textit{k} results in a higher recall at the cost of retrieving more false positives. Owing to the highly imbalanced distribution in this multi-class dataset, resultant from the low cardinality and high granularity in diagnoses present in the database, the metrics of AU-PR, $F_1$-score and RP80 were computed using a micro-averaged approach. $F_1$-score was computed using a probability threshold of 0.5 for each class, whereas AU-PR and RP80 were calculated using variable thresholds.

In spite of being a more relevant and interpretable metric for binary prediction scenarios such as readmission prediction, RP80 was also computed for diagnoses prediction models to support a direct comparison with other existing works such as clinicalBERT\_multi. However, it is important to note that in diagnoses prediction applications the importance of false positive generation can vary with the class being predicted, which further questions the importance of RP80 in this scenario as discussed more in-depth in Section~\ref{sec:discussion}.

\begin{landscape}
    \begingroup
\renewcommand{\arraystretch}{0.95}
\begin{table}[!th]
\centering
\caption[Diagnoses prediction results (ICD-9 codes) for the individual admission model architecture.]{Diagnoses prediction results (ICD-9 codes) for the individual admission model architecture. Results reported for 10-fold cross-validation. Micro values presented for AU-PR, $F_1$-score and RP80. Best values for each metric highlighted in bold.}
\label{tab:chap5:results_individual_diagnosis_icd}
\begin{tabular}{@{}lc @{}cccccc@{}}
\toprule
Features & & AU-PR & $F_1$-score & RP80 & Recall@10 & Recall@20 &Recall@30 \\
\cmidrule{1-1}\cmidrule{3-8}
Text  & & 0.241 & 0.004 & 0.001 & 0.277 & 0.417 & 0.509 \\
Diag ICD9 & & 0.364 & 0.258 & 0.086 & 0.356 & 0.496 & 0.584 \\
Text + Diag ICD9 & & 0.365 & 0.293 & 0.083 & 0.361 & 0.498 & 0.584 \\
Diag ICD9 + Days to prior admission & & 0.367 & 0.283 & \textbf{0.088} & 0.358 & 0.502 & 0.588 \\
Text + Diag ICD9 + CUI & & 0.357 & 0.254 & 0.077 & 0.346 & 0.487 & 0.576 \\
Text + Diag ICD9 + Proc ICD9  & & 0.362 & 0.262 & 0.079 & 0.351 & 0.493 & 0.582 \\
Text + Diag ICD9 + Days to prior admission & & \textbf{0.370} & \textbf{0.316} & 0.083 & \textbf{0.365} & \textbf{0.506} & \textbf{0.590} \\
Text + Diag ICD9 + Days to prior admission + Proc ICD9  & & 0.362 & 0.270 & 0.082 & 0.352 & 0.494 & 0.582 \\
Text + Diag ICD9 + Days to prior admission + Proc ICD9 + CUI & & 0.364 & 0.266 & 0.086 & 0.353 & 0.496 & 0.584 \\
\cmidrule{1-1}\cmidrule{3-8}
LIG-Doctor~\cite{LIGDoctor2020} --- Diag ICD9 & & --- & --- & --- & 0.420 & 0.580 & 0.670 \\
\bottomrule
\end{tabular}
\end{table}
\endgroup

    \begingroup
\renewcommand{\arraystretch}{0.95}
\begin{table}[!ht]
\centering
\caption[Diagnoses prediction results (CCS codes) for the individual admission model architecture.]{Diagnoses prediction results (CCS codes) for the individual admission model architecture. Results reported for 10-fold cross-validation. Micro values presented for AU-PR, $F_1$-score and RP80. Best values for each metric highlighted in bold.}
\label{tab:chap5:results_individual_diagnosis_ccs}
\begin{tabular}{@{}lc cccccc@{}}
\toprule
Features & & AU-PR & $F_1$-score & RP80 & Recall@10 & Recall@20 &Recall@30 \\
\cmidrule{1-1}\cmidrule{3-8}
Text  & & 0.281 & 0.058 & 0.001 & 0.299 & 0.471 & 0.598 \\
Diag CCS & & 0.437 & 0.320 & 0.101 & 0.417 & 0.583 & 0.686 \\
Text + Diag CCS & & \textbf{0.444} & \textbf{0.346} & 0.104 & \textbf{0.426} & \textbf{0.597} & \textbf{0.699} \\
Diag CCS + Days to prior admission & & 0.439 & 0.323 & \textbf{0.107} & 0.421 & 0.591 & 0.694 \\
Text + Diag CCS + CUI & & 0.425 & 0.299 & 0.090 & 0.406 & 0.576 & 0.680 \\
Text + Diag CCS + Proc CCS  & & 0.431 & 0.315 & 0.094 & 0.413 & 0.583 & 0.686 \\
Text + Diag CCS + Days to prior admission & & 0.426 & 0.306 & 0.091 & 0.407 & 0.579 & 0.682 \\
Text + Diag CCS + Days to prior admission + Proc CCS  & & 0.433 & 0.321 & 0.101 & 0.415 & 0.585 & 0.687 \\
Text + Diag CCS + Days to prior admission + Proc CCS + CUI & & 0.429 & 0.322 & 0.095 & 0.412 & 0.583 & 0.685 \\
\cmidrule{1-1}\cmidrule{3-8}
ClinicalBERT\_multi~\cite{franz2020DeepObserver} --- Text & & 0.408 & --- & 0.081 & --- & --- & --- \\
LIG-Doctor~\cite{LIGDoctor2020} --- Diag CCS & & --- & --- & --- & 0.520 & 0.680 & 0.760 \\
\bottomrule
\end{tabular}
\end{table}
\endgroup
\end{landscape}

Obtained results for the prediction of ICD-9 diagnoses codes are presented in Table~\ref{tab:chap5:results_individual_diagnosis_icd}. Performances from LIG-Doctor~\cite{LIGDoctor2020} (a sequence-based model) are reported for comparison. Table~\ref{tab:chap5:results_individual_diagnosis_ccs} summarises obtained results for the prediction of CCS diagnoses, and presents performances from clinicalBERT\_multi~\cite{franz2020DeepObserver} and LIG-Doctor~\cite{LIGDoctor2020} for comparison.

\subsection{Sequence Modelling}

Previous results on readmission and diagnoses prediction were obtained using the first model architecture (Figure~\ref{fig:chap5:modelarchitecture}), which explored patient admissions from an individual standpoint. However, during the process of clinical reasoning physicians can explore the medical history of a patient and use that information to adjust their decisions, for instance when defining diagnoses and procedural interventions. Taking this into consideration, a second model architecture was created that reconstructs patient trajectories using sliding windows containing a fixed number of past patient admissions, thus using subsequences of patients admissions to predict the clinical outcomes.

The intuition behind the process of scanning through a trajectory with a sliding window was that by iteratively providing the model with sequences of admissions, which had information on how the medical status progressed throughout time, the model would be capable of more efficiently learning disease progression dynamics, and therefore provide better outcome predictions. Exploring larger window sizes would let the model capture information (and relations) from a more distant past, whereas smaller windows would provide a more recent introspection into the patient medical history. However, considering the reduced size of the dataset, the use of smaller window sizes had the advantage of creating a larger number of samples, since patients with longer medical histories had their trajectories decomposed into multiple smaller subsets of hospital visits.

Since the newly proposed architecture (Figure~\ref{fig:chap5:trajectorymodelarchitecture}) leveraged the first model, which generated predictions using isolated admissions, to convert admissions inside the sliding window into dense representations that could then be fed to the recurrent network, the selection of the model to be used for converting admissions into dense representations depended on the prediction task to be addressed by the sequence-based model. As an example, a sequence-based model targeting ICD-9 code prediction should use the best performing model from Table~\ref{tab:chap5:results_individual_diagnosis_icd} for the admission conversion process.

However, the implementation of the model in Figure~\ref{fig:chap5:trajectorymodelarchitecture} required each patient admission to have a fixed dense representation. As previously explained, when clinical text was used as an input feature (in Figure~\ref{fig:chap5:modelarchitecture}) each patient admission was in fact subdivided into multiple sub-admissions, each with a fragment of the clinical note. Since the \textquotedblleft merging\textquotedblright~procedure only occurred after class probabilities were computed, it was not possible to obtain a single fixed representation for each admission when using clinical text. Therefore, models trained with clinical text could not be used in the sequence-based architecture.

Owing to the existence of related work exploring sequences of patient admissions to predict diagnoses codes, preliminary experiments with the proposed architecture focused on diagnoses prediction (CCS codes), using the \textquotedblleft Diag CCS + Days to prior admission\textquotedblright~model configuration from Table~\ref{tab:chap5:results_individual_diagnosis_ccs} to convert admissions into embedding representations. Preliminary results presented in Table~\ref{tab:chap5:results_trajectory_diagnosis_ccs} were obtained using 10-fold cross-validation. Evaluation metrics were maintained to enable a more direct comparison between the two proposed model architectures. Additionally, for comparison purposes, a set of experiments was conducted using the diagnoses input feature converted to multi-hot input format, as used by Rodrigues-Jr~\textit{et al.}~\cite{LIGDoctor2020}. These results are presented in Table~\ref{tab:chap5:results_trajectory_diagnosis_multihot_ccs} and were also obtained using 10-fold cross-validation.

\section{Discussion}
\label{sec:discussion}

In this section we present and discuss the major findings resultant from the present study, dividing them into the four main categories of initial experiments, readmission prediction, diagnoses prediction, and sequence modelling. 

\begin{landscape}
    \begingroup
\renewcommand{\arraystretch}{0.95}
\begin{table}[!th]
\centering
\caption[Preliminary results for diagnoses prediction (CCS codes) using the admission sequence-based model architecture.]{Preliminary results for diagnoses prediction (CCS codes) using the admission sequence-based model architecture. Results reported for 10-fold cross-validation. Micro values presented for AU-PR, $F_1$-score and RP80. Best values highlighted in bold.}
\label{tab:chap5:results_trajectory_diagnosis_ccs}
\begin{tabular}{cccccc cccccc}
\toprule
\begin{tabular}[c]{@{}c@{}}Layer\\Type\end{tabular} & \begin{tabular}[c]{@{}c@{}}Layer\\Size\end{tabular} & \begin{tabular}[c]{@{}c@{}}Num\\Layers\end{tabular} & Bidirectional & \begin{tabular}[c]{@{}c@{}}Sliding \\Window \end{tabular} & & AU-PR & $F_1$-score & RP80 & Recall@10 & Recall@20 &Recall@30 \\
\cmidrule{1-5}\cmidrule{7-12}

LSTM & 255 & 1 & False & 3 & & 0.109 & 0.099 & 0.001 & 0.212 & 0.353 & 0.467 \\
LSTM & 255 & 1 & True  & 3 & & 0.127 & 0.129 & 0.001 & 0.223 & 0.366 & 0.473 \\
LSTM & 818 & 1 & True  & 3 & & 0.137 & 0.156 & 0.001 & 0.206 & 0.350 & 0.462 \\
LSTM & 818 & 1 & True  & 6 & & 0.132 & 0.152 & 0.001 & 0.209 & 0.351 & 0.462 \\
LSTM & 255 & 2 & True  & 3 & & 0.134 & 0.182 & 0.001 & 0.234 & 0.399 & 0.522 \\
LSTM & 818 & 2 & True  & 3 & & \textbf{0.174} &	\textbf{0.212} & 0.001 & \textbf{0.245} & \textbf{0.408} & \textbf{0.531} \\
LSTM & 818 & 2 & True  & 6 & & 0.165 &	0.206 & 0.001 &	0.241 &	\textbf{0.408} & \textbf{0.531} \\
LSTM & 818 & 3 & True  & 3 & & 0.117 &	0.149 & 0.001 &	0.232 &	0.407 & \textbf{0.531} \\
GRU  & 818 & 2 & True  & 3 & & 0.138 &	0.204 & 0.001 &	0.224 &	0.381 & 0.502 \\
\bottomrule
\end{tabular}
\end{table}
\endgroup
   
    \begingroup
\renewcommand{\arraystretch}{0.95}
\begin{table}[!th]
\centering
\caption[Preliminary results for diagnoses prediction (CCS codes) using the admission sequence-based model architecture with multi-hot diagnoses input.]{Preliminary results for diagnoses prediction (CCS codes) using the admission sequence-based model architecture with multi-hot diagnoses input. Average results reported for 10-fold cross-validation. Micro values presented for AU-PR, $F_1$-score and RP80. Best values for each metric are highlighted in bold.}
\label{tab:chap5:results_trajectory_diagnosis_multihot_ccs}
\begin{tabular}{cccccc cccccc}
\toprule
\begin{tabular}[c]{@{}c@{}}Layer\\Type\end{tabular} & \begin{tabular}[c]{@{}c@{}}Layer\\Size\end{tabular} & \begin{tabular}[c]{@{}c@{}}Num\\Layers\end{tabular} & Bidirectional & \begin{tabular}[c]{@{}c@{}}Sliding \\Window \end{tabular} & & AU-PR & $F_1$-score & RP80 & Recall@10 & Recall@20 &Recall@30 \\
\cmidrule{1-5}\cmidrule{7-12}
GRU	 & 255 & 1 & False & 3 & & 0.466 & 0.324 & 0.123 & 0.433 & 0.605 & 0.707 \\
GRU	 & 255 & 1 & True  & 3 & & \textbf{0.503} & \textbf{0.400} & 0.153 & \textbf{0.464} & \textbf{0.638} & \textbf{0.736} \\
GRU	 & 255 & 1 & True  & 6 & & 0.486 & 0.377 & 0.137 & 0.456 & 0.627 & 0.726 \\
GRU	 & 255 & 2 & True  & 3 & & 0.492 & 0.392 & \textbf{0.157} & 0.454 & 0.627 & 0.726 \\
GRU	 & 255 & 2 & True  & 6 & & 0.478 & 0.372 & 0.142 & 0.448 & 0.619 & 0.718 \\
LSTM & 255 & 1 & True  & 3 & & 0.478 & 0.353 & 0.125 & 0.442 & 0.614 & 0.715 \\
LSTM & 255 & 1 & True  & 6 & & 0.458 & 0.319 & 0.107 & 0.431 & 0.601 & 0.704 \\
\cmidrule{1-5}\cmidrule{7-12}
\begin{tabular}[c]{@{}c@{}}MGRU\\(LIG-Doctor~\cite{LIGDoctor2020})\end{tabular} & 271 & 1 & True & -- & & --- & --- & --- & 0.520 & 0.680 & 0.760 \\
\bottomrule
\end{tabular}
\end{table}
\endgroup
\end{landscape}


\subsection{Initial Experiments}

Beginning with the first experiments on readmission prediction, which focused on exploring the proposed model architecture for isolated patient admissions, a base feature set was used comprising clinical text, admission duration and elapsed days since the patient was discharged from a prior admission. Clinical text was represented using clinicalBERT, a public model made available by Huang~\textit{et al.} in pretrained and fine-tuned variants, hence initial tests used the original dataset from~\cite{clinicalBERT2020}. To enable a more consistent assessment of model performance, the dataset was used in a 5-fold cross-validation setting.

Despite also mentioning the use of 5-fold cross-validation in model development, Huang~\textit{et al.} only made available a single fine-tuned model, which raised concerns of data leakage when applying this fine-tuned model to the same dataset. In other words, there was a high probability of having inflated test performances when using this model, as it could be evaluated in test partitions containing samples that had previously been used to train the fine-tuned model.

The top half of Table~\ref{tab:chap5:results_individual_readmission_experiments1} presents obtained results for the original dataset using the fine-tuned clinicalBERT model and 5-fold cross-validation. The top row shows the performance of the proposed model when using only clinical text, which was equivalent to using the complete model architecture from~\cite{clinicalBERT2020}. Obtained results reached a RP80 of 0.614, which was vastly higher than the 0.242 reported by the authors of clinicalBERT in the same dataset and confirmed our suspicion of data leakage. Regarding temporal features, while duration and days since the previous admission had some predictive power when used separately (RP80 of 0.025 and 0.033, respectively), the combination of both features as input significantly increased their predictive power, reaching a RP80 of 0.238 which was very close to the 0.242 originally obtained using text. For the final test, clinical text was used with both temporal features, attaining even higher performances with an average RP80 of 0.904 across all folds. Although results obtained with clinical text were deemed invalid due to the noticeable presence of information cross-talk, the combination of text with other feature modalities provided interesting insight as the model effectively managed to fuse the predictive power of all features. Furthermore, since temporal features were independent from clinicalBERT and had not been previously used in the dataset, obtained results provided an actual idea of their predictive potential for this prediction problem.

For the next tests, fine-tuned clinicalBERT was replaced with a pretrained version of the same model, and a different dataset was created that had a larger pool of negative samples available for selection. Class balancing was performed at the beginning of each cross-validation fold by randomly selecting negative samples such that a class distribution of 1:1 was reached, matching the distribution found in the original dataset~\cite{clinicalBERT2020}. Moreover, cross-validation was changed from 5 to 10 folds, using eight folds for training, one fold for validation (to train the model using early-stopping) and one fold to test the resulting model. Obtained results are presented at the bottom half of Table~\ref{tab:chap5:results_individual_readmission_experiments1}. 

The first aspect that stood out was the drastic performance drop when using clinical text as one of the input feature in this scenario. While $F_1$-score, precision and recall (all of the metrics computed with a fixed probability threshold of 0.5) showed similar values, the area under the precision-recall curve and RP80 decreased significantly. Regarding temporal features, duration showed a slight decrease in AU-PR and RP80 whereas days to prior admission showed the opposite, with some improvement in the same metrics. Similarly to what was observed in the first scenario, the combination of both temporal features provided further performance improvements (RP80 increased to 0.358).

\subsection{Readmission Prediction}

Previous experiments provided important knowledge on the impact of each feature in readmission prediction, as well as on how the proposed model handled different feature combinations. However, the datasets explored so far were based on the assumption that admissions from the same patient were independent, which we consider fundamentally incorrect.
Therefore, for the final results on readmission prediction with the first architecture, a third dataset was created where a constraint was placed on the distribution of patient information across folds,~\textit{i.e.} all admissions from a patient had to be enclosed in single dataset fold.

In Table~\ref{tab:chap5:results_individual_readmission} one can observe the final results, obtained using 10-fold cross-validation. Clinical text maintained the same trend as before, with similar values on $F_1$-score, precision, recall and AUC but low AU-PR and RP80, whilst admission duration demonstrated slight improvements in most metrics. However, days to prior admission showed significant improvement in RP80 with an increase from 0.082 to 0.373 (50\% increase over the 0.242 reported by clinicalBERT). Moreover, this increase in performance was further accentuated through the addition of admission duration with RP80 reaching a value of 0.481, demonstrating a model with higher capacity of assigning less false positives. It was also interesting to notice that the combination of clinical text with these features led to an increase in precision and AUC, but penalized the remaining metrics. The impact of additional features such as diagnoses codes was also assessed, leading to a drop in performance across all evaluation metrics. 

Despite the below-par performances of clinical text when using pretrained clinicalBERT, its usage as an input feature for readmission prediction revealed three interesting trends that should be retained for further experimentation: 1) when using pretrained clinicalBERT, clinical text led to an improvement in precision but at the cost of reducing recall by a higher margin; 2) when using fine-tuned clinicalBERT to represent clinical text, text can achieve interesting performances as demonstrated in~\cite{clinicalBERT2020}; and 3) the positive reinforcement from combining fine-tuned clinicalBERT with temporal features (as observed in Table~\ref{tab:chap5:results_individual_readmission_experiments1}) brings interesting prospects for the results in Table~\ref{tab:chap5:results_individual_readmission} through the usage of fine-tuned clinicalBERT models to represent clinical text, instead of the pretrained model that was used in this work.

\subsection{Diagnoses Prediction}

For the second prediction task, it was possible to maintain the same model architecture by addressing the multi-label multi-class problem of diagnoses prediction as multiple binary classification problems. In fact, the process of letting the model predict the probability of each diagnosis code in an individual manner enabled the correct modelling of an actual diagnosing process, where a patient can be diagnosed with multiple co-occurring diseases, contrary to what occurs in normal multi-class problems where outcomes are predicted in a mutually exclusive manner (in a sort of one-disease-versus-all).

The proposed architecture was used for two variations of diagnoses prediction: one targeting the prediction of ICD-9 codes (Table~\ref{tab:chap5:results_individual_diagnosis_icd}) and the other targeting CCS codes (Table~\ref{tab:chap5:results_individual_diagnosis_ccs}). Model development followed the 10-fold cross-validation strategy using the diagnoses prediction dataset without any sampling mechanism, contrary to other related works~\cite{franz2020DeepObserver}.

Due to the different nature of this problem new evaluation metrics were introduced, namely Recall@\textit{k} which assessed model capability to retrieve all the correct diagnoses codes in a list containing the top~\textit{k} diagnostic code predictions. Still regarding evaluation metrics, it should be noted that in diagnoses prediction applications the importance of false positive generation can greatly vary with respect to the class being predicted. While micro-averaged RP80 does not consider the importance of each class equally, contrary to the macro-averaged variant which effectively does that, it still fails to capture the varying importance of false positives for each diagnosis class as a false positive for a less common disease is very different from a false positive for a common one. For instance, a false positive for a disease such as brain cancer is more important than a false negative as it can trigger further screening on the patient, whereas for diseases like the flu false positives have less importance. Nonetheless, micro RP80 was computed for each model to enable a direct performance comparison with clinicalBERT\_multi~\cite{franz2020DeepObserver}.

Beginning with the results for diagnoses prediction with ICD-9 as the prediction target (Table~\ref{tab:chap5:results_individual_diagnosis_icd}), although using clinical text alone resulted in the lowest predictive performance among all combinations, the fact that its Recall@30 reached a value of 0.509 is an interesting result considering that we used a pretrained version of clinicalBERT instead of a fine-tuned one. Regarding diagnostic information, its isolated use attained a micro RP80 of 0.086 and a micro $F_1$ of 0.258. However, the combination of clinical text with diagnostic information led to slight performance improvements, with the highest improvement being in micro $F_1$ which increased from 0.258 to 0.293. However, it also led to a small reduction in micro RP80. The addition of elapsed days until the last admission to the feature subset further improved model performance, resulting in the best performing configuration except for the micro RP80 metric. The inclusion of additional features such as procedure and medication information did not bring further performance gains, resulting in fact a small performance loss. The bottom row in Table~\ref{tab:chap5:results_individual_diagnosis_icd} presents the performance of LIG-Doctor when predicting ICD-9 codes for the next patient admission. These results are merely indicative due to several factors: 1) LIG-Doctor uses sequences of admissions instead of isolated admissions, thus being more directly comparable with our second proposed architecture, 2) it encodes diagnostic information differently at the input layer, using multi-hot encoding, 3) it uses a simplified search space containing only the ICD-9 present in MIMIC-III, reducing its dimension to 855 (30\% smaller than our ICD-9 search space) and making it less generalizable for possible future samples, and 4) due to the model development strategy followed by the authors, which will be discussed in more detail in Section~\ref{discussion:sequence}.

The next step involved testing the same model architecture but for the prediction of CCS codes, with the expectation of obtaining better results due to the smaller search space for prediction (Table~\ref{tab:chap5:results_individual_diagnosis_ccs}). Clinical text and diagnosis information exhibited a similar trend as before, but with better performances due to the smaller size of CCS. The combination of both features presented once again an increase in predictive power, obtaining the top performance in most metrics and reaching a Recall@30 of approximately 70\%. The addition of temporal information to diagnostic inputs demonstrated a reduced increase in RP80 ($\uparrow$ 0.03) at the cost of a slight reduction in the remaining metrics comparatively to the top performing combination. Nevertheless, a different behaviour was seen for the fusion of text, diagnosis information and other features (either temporal or coding features), which showed a performance decay across all metrics.

The two bottom rows in Table~\ref{tab:chap5:results_individual_diagnosis_ccs} present two different solutions on CCS code classification. Despite now using a complete search space for the prediction target, LIG-Doctor is still presented only for illustrative purposes due to its model architecture being sequence-oriented. Regarding clinicalBERT\_multi, it is also not possible to perform an actual direct comparison as this solution did not perform diagnoses prediction for future visits but instead attempted to detect the codes for the present patient admission,~\textit{i.e.} given a clinical note from an admission the system predicted diagnoses codes for the same admission, and the authors used a sampling procedure to ensure that the same label distribution existed in the train-val-test splits. Nonetheless, documented results are very interesting considering that the authors only used fine-tuned clinicalBERT and clinical notes for diagnoses classification, which similarly to what was verified in readmission prediction brings us future prospects on the experimentation with the fine-tuned clinicalBERT for leveraging clinical notes as an input feature.

\subsection{Sequence Modelling}
\label{discussion:sequence}

The second model architecture herein proposed (Figure~\ref{fig:chap5:trajectorymodelarchitecture}) had the objective of augmenting the first one by leveraging it in a recurrent-based architecture capable of handling sequences of patient admissions, instead of exploring patient admissions as isolated views. The reasoning behind such approach was that by providing access to information on the progression of the patient health status, the solution would have the possibility of more accurately capturing and modelling disease dynamics, which could in turn result in better outcome predictions. Similar concepts have already been explored in related works such as DeepCare~\cite{pham2017DeepCare}, AdaptiveNet~\cite{AdaptiveNet2020} and LIG-Doctor~\cite{LIGDoctor2020}.

To use the proposed architecture, it was firstly necessary to define a scheme for structuring the sequences of patient admissions before they could be used as model input. Similar existing solutions such as LIG-Doctor have accomplished this by assembling and storing the longest possible sequence for each patient, and iterating over it to generate child subsequences (\textit{e.g.} a sequence of admissions $A-B-C\rightarrow D$ would originate the following child sequences: $A-B\rightarrow  C$ and $A\rightarrow  B$). However, we created a different approach based on sliding windows which scanned through patient trajectories and stored the corresponding subsequences of admissions. As an example, given a window of size 2 and the sequence of admissions $A-B-C\rightarrow D$, the system would process the following subsequences: $A-B\rightarrow C$ and $B-C\rightarrow D$. By adjusting the size of the sliding window, it was possible to define the amount of contextual information the model could attend to. The objective of this approach was to develop models capable of learning sub-patterns/dynamics of disease progression, instead of memorising larger trajectories by having access to the full trajectory at once.

Table~\ref{tab:chap5:results_trajectory_diagnosis_ccs} presents a list of preliminary results for diagnoses prediction (CCS codes) obtained with the sequence-based model. Since this model was based on recurrent networks, the first factor to be evaluated was the impact of bidirectionality. The first two rows in this table present obtained results for a similar setting where only bidirectionality was varied, demonstrating performance improvements when using bidirectionality with more substantial gains being verified for micro AU-PR and micro $F_1$-score. Next, layer size was increased from 255, the dimension of the prediction search space, to 818 which was the size of the admission dense representation extracted using the first model architecture. Obtained results showed a trade-off where increasing layer size led to an improvement in precision (gains in AU-PR and $F_1$) and a reduction in recall (visible in the all levels of Recall@\textit{k}). The same model configuration was tested with a larger sliding window (increased from 3 to 6 admissions) but no significant impact was verified in obtained results.

The following parameter to be evaluated was the number of layers in the recurrent network, where the addition of an extra recurrent layer resulted in noticeable performance gains when using either layer size (255 or 818). However, further adding a third recurrent layer had the opposite result, reducing model precision (visible in the reduced AU-PR and $F_1$) whereas recall remained relatively unchanged. For the final test with the LSTM network architecture, the best performing configuration (2 layers with size 818)
was evaluated with sliding windows of 6 admissions. The increase in contextual information being forwarded to the model led to a small reduction in model precision. Finally, the impact of network typology was assessed by using the best parameter configuration with GRU instead of LSTM networks. Obtained results are presented in the bottom row of Table~\ref{tab:chap5:results_trajectory_diagnosis_ccs} and show an overall performance reduction from using a different architecture.

On the one hand, a comparison between prediction performance when using separate admissions (Table~\ref{tab:chap5:results_individual_diagnosis_ccs}) and sequences of admissions (Table~\ref{tab:chap5:results_trajectory_diagnosis_ccs}) revealed a significant decrease in performance when using sequences of admissions, which consisted in the opposite behaviour we expected. For instance, Recall@\textit{k} saw performance drops between 16 and 18 percentage points, $F_1$ and RP80 decreased approximately 10 percentage points, and AU-PR fell nearly 26 percentage points. On the other hand, the authors of LIG-Doctor documented a Recall@30 of 0.760 when using admission sequences as model input, which proved the potential of leveraging more information from the patient trajectories in this prediction problem. It should be noted that we do not compare our results with DeepCare as it used different datasets, metrics and prediction outcomes (DeepCare simplifies the multi-label multi-class diagnoses prediction problem in two completely separate binary problems), nor with AdaptiveNet since it also used different datasets and metrics, and consisted in a regression problem.

LIG-Doctor was similar to our proposed architecture, differing mostly at the input encoding stage where it encoded diagnoses codes with a multi-hot approach, and used the resulting multi-hot vectors as input. However, its performances should not be used for a head-to-head comparison due to certain implementation nuances, namely the selected activation function and the strategy used for model development. Regarding the first aspect, LIG-Doctor used Softmax to compute class predictions, which as previously mentioned does not correctly capture the problem of diagnoses modelling since it computes probabilities based on a mutual-exclusion property. Concerning the second aspect, their model development strategy only used three random runs where samples had the possibility of never being used for training or evaluation (\textit{e.g.} the same sample could be assigned 3 times to the training dataset and never be evaluated on). Furthermore, it only comprehended training and validation splits. The validation partition was used to train the model using early-stopping, with the best performing model not being tested on a separate test partition thus yielding more optimistic performances.

To have a more stable baseline for comparison, we followed the same principle as LIG-Doctor and converted diagnoses codes into multi-hot vectors. By assessing the impact of using this input feature instead of the admission representations extracted using the model from Figure~\ref{fig:chap5:modelarchitecture}, one could in theory determine the importance of input encoding in system performance. The newly encoded input was used in the proposed architecture (Figure~\ref{fig:chap5:trajectorymodelarchitecture}), where Sigmoid activation was used to compute class probabilities. Model development followed the same 10-fold cross-validation approach as before.

Obtained results (Table~\ref{tab:chap5:results_trajectory_diagnosis_multihot_ccs}) were lower than original ones (last row of Table~\ref{tab:chap5:results_trajectory_diagnosis_multihot_ccs}), with the best performing configuration attaining Recall@\textit{k} values between 2 to 6 percentage points lower. The use of bidirectionality and a smaller number of layers led to better performing systems, as described in~\cite{LIGDoctor2020}, whereas increasing sliding window size resulted in performance drops. It is possible to see that the LSTM network had worse predictive performance than GRU, contrary to what we observed in Table~\ref{tab:chap5:results_trajectory_diagnosis_ccs} where the LSTM and additional layer (layer size of 2) were present in the top performing configuration.

The abovementioned baseline served for twofold: 1) corroborate the potential in exploring the patient trajectory for prediction instead of limiting models to isolated views on the data (\textit{e.g.} individual admissions); and 2) pinpoint the key source for the verified performance loss in the proposed architecture, which was related with input encoding.
Although the first architecture managed good performances with the bottlenecking procedure, where the representation of all diagnoses codes was condensed into a single embedding of dimension 768, the second architecture struggled with the same representation as it could be losing its discriminative power during the bottlenecking process. A potential cause for such behaviour is the fact that in the second architecture all admission representations were fixed and precomputed, whereas in the first architecture the dense layer responsible for the bottlenecking procedure had the possibility of adjusting itself (and the resulting representation) during model training, hence leading to better results.

\section{Conclusion}
\label{sec:Conclusion}

The present work aimed to explore recent techniques for the representation and merging of multimodal patient information, as well as the process of modelling patient trajectories for the prediction of different relevant clinical outcomes. Herein, we proposed two DL-based architectures that explored admissions from patient trajectories on an individual and sequence basis.

The first architecture leveraged clinical text with clinicalBERT embeddings, temporal features with dense layers, and coding features with SapBERT embeddings combined with dense layers responsible for reducing representation size. 
The resulting model provided flexibility in feature selection, enabling the use of any feature combination in model development, and was used for readmission prediction and diagnoses prediction. The latter explored two different target vocabularies, a simplified version of ICD-9 and CCS, with CCS having an approximately five times smaller search space.

The second architecture integrated the first one for the extraction of dense admission representations, and used subsets of ordered admission representations to predict clinical outcomes. To capture the full span of patient trajectories, a sliding window mechanism was created that scanned through the patient trajectory feeding the model with smaller sequences of patient admissions (\textit{e.g.} windows of 3 or 6 admissions). The intuition behind this iterative scheme was that it could help the system learning the underlying dynamics of patient progression, and thus more accurately model patient trajectories. Preliminary tests with this architecture were focused on the prediction of CCS codes.

Obtained results demonstrated the potential of the first architecture to model readmission and diagnoses prediction using single patient admissions. The fact that a pretrained version of clinicalBERT was used in this architecture negatively impacted on the discriminative power of clinical text as an input feature, leading to performances far from those of clinicalBERT~\cite{clinicalBERT2020} and clinicalBERT\_multi~\cite{franz2020DeepObserver}. Nonetheless, the combination of clinical text with other types of information led to interesting findings, raising future work prospects regarding the inclusion of a fine-tuned version of clinicalBERT with the expectation of further improving presented results.

Concerning the sequence-based architecture, preliminary tests showed a decrease in predictive performance due to the selected representation for input data. We believe this behaviour was related with precomputed admission representations being used in this model. Since the model could not optimize the bottlenecking procedure responsible for generating the representation of input diagnoses, this feature lost information and discriminative power thus impacting on the admission representation and end system performance. As future work, one could experiment with a fully trainable version of this architecture, where admission representations can be adjusted during training. A different encoding for diagnoses input could also be explored by combining multi-hot encoding with SapBERT embeddings. 
Although we did not explore clinical text in these preliminary tests, we intend to do so in the future. For that, future work should be conducted with clinicalBERT in the first architecture to reevaluate the discriminative potential of clinical text and of its merging with other features, and a mechanism should be developed to aggregate the representations of multiple text chunks in a single clinical note representation.

Other possible future lines of work involve using the proposed architectures for other prediction problems, such as predicting procedure codes or medication changes. For the former, despite having a smaller search space (100 codes in the simplified version of ICD-9), one must also consider that this further reduces dataset size since there exist less admissions in MIMIC-III with this type of information. For the latter, it is important to consider its much larger search space which is in the order of 20~000 possible codes.


The proposed system is publicly available and can be currently accessed at \url{https://github.com/bioinformatics-ua/PatientTM}.

\section{Acknowledgements}
Jo\~ao Figueira Silva is funded by the FCT - Foundation for Science and Technology (national funds) under the grant PD/BD/142878/2018.

\bibliographystyle{elsarticle-num}
\bibliography{references.bib}

\begin{landscape}
    \appendix
    \setcounter{table}{0} 
    \section{Dataset composition using 10-fold stratified distribution at the patient level.}
    \begingroup
\renewcommand{\arraystretch}{0.9}
\begin{table}[!h]
\centering
\caption[Dataset distributions for the readmission prediction and diagnosis prediction tasks.]{Dataset distributions for the readmission prediction and diagnosis prediction tasks, obtained using stratified distribution at the patient level. Values reported in Fold columns correspond to the number of admissions present in each fold.}
\label{tab:chap5:hadm_distribution}
\begin{tabular}{cc@{}c@{}cccccccccc@{}c@{}}
\toprule
&  &  &  & \multicolumn{10}{c}{Fold} \\
Task & & Label & & 1 & 2 & 3 & 4 & 5 & 6 & 7 & 8 & 9 & 10 \\
\cmidrule{1-1}\cmidrule{3-3}\cmidrule{5-14}
\multirow{2}{*}{Readmission} & & 0 & & 4128 & 4228 & 4018 & 4091 & 4136 & 4124 & 4079 & 4049 & 4053 & 4066 \\
& & 1 & & 280 & 323 & 264 & 285 & 281 & 297 & 276 & 306 & 288 & 307 \\
\cmidrule{3-3}\cmidrule{5-14} 
Diagnosis & & --- & & \multicolumn{1}{l}{1073} & \multicolumn{1}{l}{1200} & \multicolumn{1}{l}{939} & \multicolumn{1}{l}{1014} & \multicolumn{1}{l}{1064} & \multicolumn{1}{l}{1079} & \multicolumn{1}{l}{990} & \multicolumn{1}{l}{1011} & \multicolumn{1}{l}{990} & \multicolumn{1}{l}{1016} \\
\bottomrule
\end{tabular}
\end{table}

\endgroup
    \setcounter{table}{0} 
    \section{Dataset composition using 10-fold stratified distribution at the admission level.}
    \begingroup
\renewcommand{\arraystretch}{0.9}
\begin{table}[!h]
\centering
\caption[Dataset distribution for the readmission prediction task, with stratified distribution at the admission level.]{Dataset distribution for the readmission prediction task, with stratified distribution at the admission level. Values reported in Fold columns correspond to the number of admissions present in each fold.}
\label{tab:chap5:hadm_distribution_hadmsplit}
\begin{tabular}{cc@{}c@{}ccccccccccc@{}}
\toprule
&  &  &  & \multicolumn{10}{c}{Fold} \\
Task & & Label & & 1 & 2 & 3 & 4 & 5 & 6 & 7 & 8 & 9 & 10 \\
\cmidrule{1-1}\cmidrule{3-3}\cmidrule{5-14}
\multirow{2}{*}{Readmission} & & 0 & & 4091 & 4106 & 4086 & 4106 & 4083 & 4123	& 4093 & 4099 & 4090 & 4095 \\
& & 1 & & 291 &  292 &  291 &  292 &  291 &  289	&  291 &  289 &  291 &  290 \\
\bottomrule
\end{tabular}
\end{table}

\endgroup

\end{landscape}

\end{document}